\documentclass[journal]{IEEEtran}
\usepackage{amsmath,graphicx,multirow,booktabs,bm,amssymb}
\ifCLASSINFOpdf
\else
\fi
\begin{document}
%
\title{Adversarial Transfer Learning for Punctuation Restoration}
%
%
%

\author{Jiangyan Yi,
        Jianhua Tao,
        Ye Bai,
        Zhengkun Tian,
        Cunhang Fan,
\thanks{The authors are with the National Laboratory of Pattern Recognition, Institute of Automation, Chinese Academy of Sciences, and University of Chinese Academy of Sciences, Beijing 100190, China (e-mail: jiangyan.yi@nlpr.ia.ac.cn, jhtao@nlpr.ia.ac.cn, ye.bai@nlpr.ia.ac.cn, zhengkun.tian@nlpr.ia.ac.cn, cunhang.fan@nlpr.ia.ac.cn). }
}

\maketitle

\begin{abstract}
Previous studies demonstrate that word embeddings and part-of-speech (POS) tags are helpful for punctuation restoration tasks.
However, two drawbacks still exist. One is that word embeddings are pre-trained by unidirectional language modeling objectives. Thus the word embeddings only contain left-to-right context information. The other is that POS tags are provided by an external POS tagger. So computation cost will be increased and incorrect predicted tags may affect the performance of restoring punctuation marks during decoding. This paper proposes adversarial transfer learning to address these problems. A pre-trained bidirectional encoder representations from transformers (BERT) model is used to initialize a punctuation model. Thus the transferred model parameters carry both left-to-right and right-to-left representations. Furthermore, adversarial multi-task learning is introduced to learn task invariant knowledge for punctuation prediction. We use an extra POS tagging task to help the training of the punctuation predicting task. Adversarial training is utilized to prevent the shared parameters from containing task specific information. We only use the punctuation predicting task to restore marks during decoding stage. Therefore, it will not need extra computation and not introduce incorrect tags from the POS tagger. Experiments are conducted on IWSLT2011 datasets. The results show that the punctuation predicting models trained with transferred parameters from pre-trained BERT model obtain significant performance gains over the models with random initialization by up to 9.4\% absolute overall $F_1$-score on test set. The results also demonstrate that the punctuation predicting models obtain further performance improvement with task invariant knowledge from the POS tagging task. Our best model outperforms the previous state-of-the-art model trained only with lexical features by up to 9.2 \% absolute overall $F_1$-score on test set.
\end{abstract}


%
\IEEEpeerreviewmaketitle

\section{Introduction}

%
Generally, the output sequences of automatic speech recognition (ASR) systems don't contain punctuation marks. Thus it degrades the readability of the generated words and leads to poor user experiences in real-world scenarios \cite{Ueffing2013Improved}. So it is necessary to restore punctuation marks for speech transcripts.

Many attempts have been made to predict punctuation marks automatically. These approaches can be roughly divided into three categories in terms of applied features: prosody features, lexical features and the combination of the previous two features based methods.

Prosody features based methods are tried by some previous researchers \cite{Christensen2001Punctuation, Kim2003A}. Christensen et al. \cite{Christensen2001Punctuation} use hidden Markov models to restore punctuation marks using acoustic data.
Kim et al. \cite{Kim2003A} try to perform punctuation prediction and speech recognition jointly with prosody features.
The previous results show that prosody features are useful, but they don't work well when speakers make pauses in unnatural places.

The combination of prosody and lexical features based methods are proposed to resolve this problem \cite{Szaszak2019Lever, Nanchen2019Empi}. Che et al. \cite{Che2016Sentence} propose to train deep neural networks (DNN) on parallel lexical and acoustic features. Tilk et al. \cite{Tilk2015LSTM} use a long short-term memory (LSTM) based punctuation prediction model trained with text and speech data by two stages. Klejch et al. \cite{Ond2016Punctuated, Ond2017Sequence} propose a recurrent neural network (RNN) encoder-decoder architecture with an attention layer to restore punctuation marks by fusing lexical and prosody features. However, these models need to utilize the lexical data with the corresponding speech data. So the use of text and speech data is limited. Yi et al. \cite{Yi2019Self} propose to train self-attention based models using word and speech embeddings. This method can use any kind of text and speech data. It also obtains obvious performance improvement with pre-trained vectors. However, it still has a limitation to utilize enough information from the text data.

In fact, it is not difficult to obtain a large amount of available text data. Therefore, this paper only focuses on lexical features based methods. A lot of studies have been tried to restore punctuation marks only using text data.

One kind of methods is that punctuation marks are treated as hidden inter-word events \cite{Liu2006Enriching}. Beeferman et al. \cite{Beeferman1998Cyberpunc} propose to train an n-gram language model (LM) using punctuated text data. The n-gram LM is also used to predict punctuation marks and perform capitalization jointly by Gravano et al. \cite{Gravano2009Restoring}.

The other kind of methods is that predicting punctuation is viewed as a sequence labeling task \cite{Ueffing2013Improved,z2018Punctuation}, in which a punctuation mark is assigned to each word.
Previous studies \cite{Wei2010Better, Ueffing2013Improved,Hasan2015Noise} show that conditional random fields (CRFs) are better-suited to predict punctuation marks than n-gram LM based methods. Lu et al. \cite{Wei2010Better} try to train CRF based models only with token features. Ueffing et al. \cite{Ueffing2013Improved} propose to combine syntactic features with LM scores, token features and sentence length to predict punctuation marks using CRF models. Part-of-speech (POS) tags of several continuous words are used as the features to train CRF and DNN combined lexical model by Cho et al. \cite{Cho2015Combin}. The results show that POS tags are helpful for improving the performance of punctuation prediction tasks.
Recently, neural networks based models are used to predict punctuation marks. Unlike the previous lexical features including n-grams, LM statics, token, POS tags and other syntax information etc., the lexical features of the neural networks are pre-trained word vectors. Che et al. \cite{Che2016Punc} propose to train DNN and convolution neural network (CNN) based models using word embeddings. The results show that the neural network based methods outperform the CRF based method over purely text data. More recently, Tilk et al. use bidirectional recurrent neural network with attention mechanism (T-BRNN) \cite{Tilk2016BRNN} to improve the performance. Yi et al. \cite{Yi2017Distilling} propose to use bidirectional LSTM (BLSTM) with a CRF layer (BLSTM-CRF) and an ensemble of models to predict punctuation. Most recently, Kim \cite{Kim2019Deep} uses deep recurrent neural networks with layer-wise multi-head attentions for punctuation restoration. The best model in \cite{Kim2019Deep} has achieved the state-of-the-art performance with purely lexical features on IWSLT2011 datasets \cite{Che2016Punc}. The overall $F_1$-score of the model is up to 68.6\%.

The aforementioned methods show that any kind of text data can be utilized through pre-trained word vectors. They also demonstrate that POS tags are useful lexical features. However, they still have two limitations.
(1) One is the word vectors are trained using left-to-right language modeling objective functions \cite{Mnih2008A, Mikolov2013}. Thus the word embeddings only have unidirectional knowledge.
(2) The other is that an extra POS tagger is needed to provide tags information for the input sequence during predicting stage. So it not only increases computation cost, but also introduces some errors from the POS tagger.
Therefore, this paper proposes adversarial transfer learning to alleviate these problems.

\begin{figure}[htb]
\hfill
\begin{minipage}[b]{1.0\linewidth}
  \centering
  \centerline{\includegraphics[width=6.0cm,height=10.5cm]{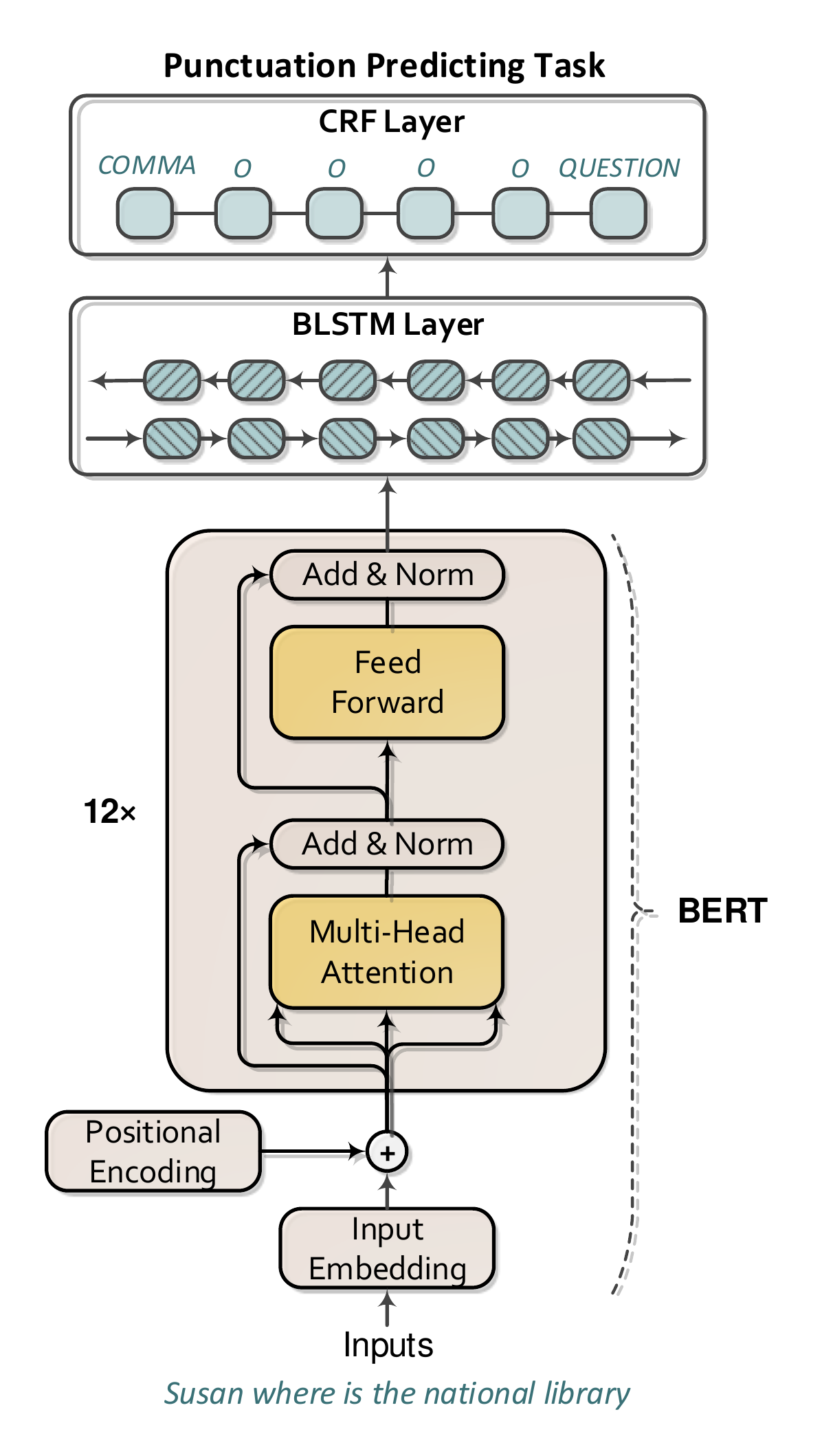}}
\end{minipage}
\caption{The architecture of BERT-BLSTM-CRF model. BERT layers are initialized by a pre-trained language representation model. BLSTM-CRF layers are initialized randomly.}
\label{fig:bert-model}
\end{figure}

Inspired by the promising results of pre-trained bidirectional encoder representations from transformers (BERT) model on many natural language processing (NLP) tasks \cite{Devlin2018BERT}, this paper tries to transfer model parameters from a pre-trained BERT model to initialize a punctuation prediction model as shown in Fig. \ref{fig:bert-model}. The BERT model is trained by fusing context from both left and right directions. Unlike word embeddings, the transferred parameters contain both left-to-right and right-to-left representations.

\begin{figure*}[htb]
\hfill
\begin{minipage}[b]{1.0\linewidth}
  \centering
  \centerline{\includegraphics[width=14.5cm,height=11.5cm]{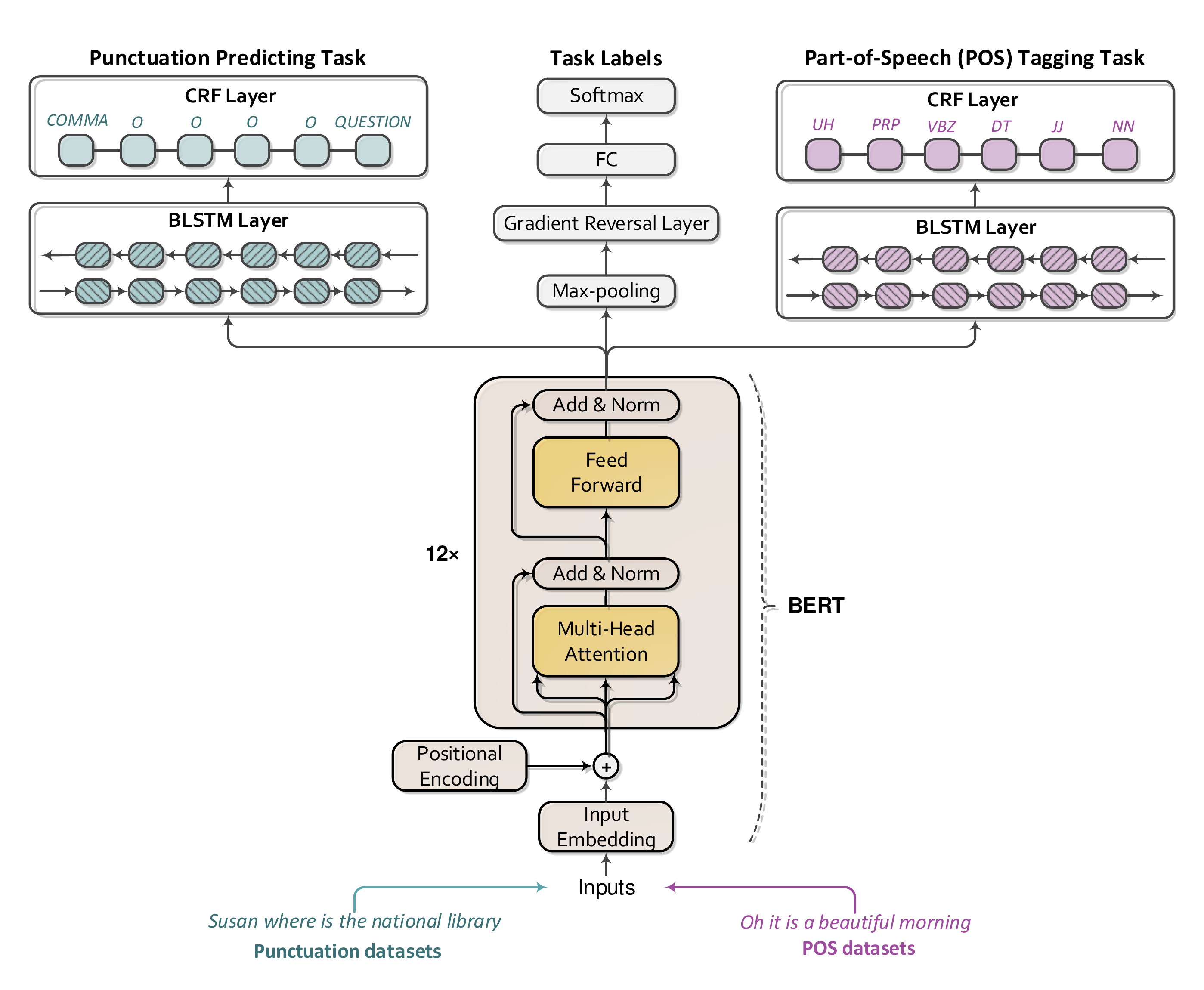}}
\end{minipage}
\caption{The architecture of the proposed adversarial BERT-BLSTM-CRF model. The task shared layers are from the pre-trained BERT model, which has a stack of 12 identical layers. The task specific classifiers are used for a punctuation predicting task and a POS tagging task, respectively. Both of them consist of BLSTM-CRF layers. FC denotes the fully connected layer. The gradient reversal layer (GRL) is introduced to ensure the feature distributions over all the tasks are as indistinguishable as possible for the task discriminator. The outputs of the task discriminator are task labels: \textit{PUN} and \textit{POS}. \textit{PUN} denotes the punctuation predicting task. \textit{POS} is referred as to the POS tagging task.}
\label{fig:adv-bert-model}
\end{figure*}

Furthermore, motivated by the success of adversarial learning \cite{Goodfellow2014Generative} on domain adaptation \cite{Ganin2017Domain}, Chinese word segmentation \cite{Chen2017Adversarial}, environment and speaker adaptation \cite{Shinohara2016Adversarial, Saon2017English} and low resource speech recognition \cite{Yi2018Adversarial, Yi2019LanguageCTC, Yi2019Language} tasks, this paper proposes to combine multi-task learning and adversarial training to solve the second limitation as shown in Fig. \ref{fig:adv-bert-model}. Multi-task learning \cite{Caruana1997Multi} a special instance of transfer learning. The conclusions drawn by Caruana in \cite{Caruana1997Multi} show that multi-task learning is effective for improving the performance of a single task, due to the extra information contained in the training signals for the other related tasks. Therefore, a POS tagging task is used as an auxiliary task to improve the performance of the punctuation prediction task. The model of the punctuation prediction task and the POS tagging task consists of shared and private layers. The shared layers contain task independent information. The task specific features are learned from the private layers of each task.
However, the shared layers may learn some unnecessary task specific information. Thus adversarial learning is used to ensure that the shared layers of the model learn more task invariant knowledge. We only use the punctuation prediction task to output punctuation marks during decoding. Thus this method can use the syntactic features from the POS tagging task without increasing any extra computation and introducing unnecessary errors from the POS tagger.

There has been no work, to the best of our knowledge, that combines transfer learning and adversarial strategy to improve the performance of punctuation restoration tasks.
The main contributions of this paper are as follows. (1) A pre-trained BERT model is used to transfer bidirectional representations to punctuation prediction models. (2) Adversarial multi-task learning is used to learn task invariant information with an extra POS tagging task for the punctuation prediction task.
Experiments are conducted on IWSLT2011 datasets. The results demonstrate that the punctuation predicting models initialized by the pre-trained BERT model obtain significant performance improvement against the models initialized randomly by up to 9.4\% absolute overall $F_1$-score on test set. The results also show that the punctuation predicting models obtain further performance gains with task invariant knowledge from the POS tagging task. Our best model achieves better results than the previous state-of-the-art model trained with purely text data \cite{Kim2019Deep} and combination of lexical and acoustic features \cite{Yi2019Self} by up to 9.2 \% and 4.9 \% absolute overall $F_1$-score on test set, respectively.

The rest of this paper is organized as follows. Section \uppercase\expandafter{\romannumeral2} briefly introduces how to transfer parameters from a pre-trained BERT model to a punctuation prediction model. How to transfer task invariant knowledge from a POS tagging task is presented in Section \uppercase\expandafter{\romannumeral3}. Experiments are described in Section \uppercase\expandafter{\romannumeral4}. The results are discussed in Section \uppercase\expandafter{\romannumeral5}. The conclusions are drawn in Section \uppercase\expandafter{\romannumeral6}.

\begin{table}
\caption{An example of inputs and outputs for the punctuation predicting task.}
\label{tab:example-pun}
\centering
\begin{tabular}{c|cccccc}
\hline
{Input words} & \textit{Susan} &  \textit{where} & \textit{is} & \textit{the} & \textit{national} & \textit{library} \\
{Output labels}& \textit{COMMA} &  \textit{O} & \textit{O} & \textit{O} & \textit{O} & \textit{QUESTION}\\
\hline
\end{tabular}
\end{table}

\begin{table}
\caption{An example of inputs and outputs for the POS tagging task.}
\label{tab:example-pos}
\centering
\begin{tabular}{c|cccccc}
\hline
{Input words} & \textit{Oh} &  \textit{it} & \textit{is} & \textit{a} & \textit{beautiful} & \textit{morning} \\
{Output labels}& \textit{UH} &  \textit{PRP} & \textit{VBZ} & \textit{DT} & \textit{JJ} & \textit{NN}\\
\hline
\end{tabular}
\end{table}

\section{Transfer parameters from Pre-trained model}

Inspired by the state-of-the-art results of pre-trained BERT on many NLP tasks [24], we initialize a punctuation prediction model by the parameters from a pre-trained BERT model. The BERT model is trained with bidirectional context information. Unlike word embeddings, the transferred parameters contain both left-to-right and right-to-left information.
The model architecture used to predict punctuation marks is shown in Fig. \ref{fig:bert-model}. It consists of BERT and BLSTM-CRF layers. The BERT layers are from a pre-trained BERT model proposed by Devlin et al. \cite{Devlin2018BERT}. The BLSTM-CRF layers are motivated by the work \cite{Huang2015Bid}. Thus the model for punctuation prediction is called BERT-BLSTM-CRF.

In this paper, predicting punctuation is viewed as a sequence labeling task. An example of inputs and outputs for the punctuation predicting task are listed in Table \ref{tab:example-pun}. The inputs are unpunctuated words, e.g. ``\textit{Susan where is the national library}''. The corresponding outputs are punctuation marks, such as ``\textit{COMMA O O O O QUESTION}''. The details of punctuation marks are described in Section \ref{exp:iwslt}.

\subsection{BERT layers}

BERT is designed to pre-train deep bidirectional representations from unlabeled text by jointly conditioning on both left and right context in all layers. The pre-trained BERT model can be used to finetune with just one additional output layer for many tasks, such as named entity recognition and question answering etc. It achieves state-of-the-art results on several NLP tasks \cite{Devlin2018BERT}.
The architecture of a BERT model is a multi-layer bidirectional Transformer encoder proposed by Vaswani et al. \cite{Vaswani2017Attention}.

The encoder consists of a stack of $N$ identical layers as shown at the bottom of Fig. \ref{fig:bert-model}. Each layer has two sub-layers. The first is a multi-head self-attention mechanism. The second is a fully connected feed-forward network. A residual connection is employed around each of the two sub-layers, followed by layer normalization.

Positional encodings are utilized to make use of the order of the input or output sequence.
The input embeddings are learnt from input tokens similarly to other sequence transduction models. The dimension of the embeddings is denoted by $d_{model}$.

An attention function can be described as mapping a query and a set of key-value pairs to an output, where the query, keys, values, and output are all vectors. Instead of performing a single attention function with $d_{model}$-dimensional keys, values and queries, Vaswani et al. find it beneficial to linearly project the queries, keys and values $h$ times with different, learned linear projections to ${d_k}$, ${d_k}$ and ${d_v}$ dimensions, respectively. Multi-head attention allows these projected versions of queries, keys and values to perform the attention function in parallel, yielding ${d_v}$-dimensional output values. Please see more details in \cite{Vaswani2017Attention, Devlin2018BERT}.

\subsection{BLSTM-CRF layers}

Motivated by the work in \cite{Huang2015Bid}, a BLSTM layer and a CRF layer are combined to form BLSTM-CRF layers as shown at the top of Fig. \ref{fig:bert-model}. The two layers can efficiently use past and future input features via the BLSTM layer and sentence level tag information via the CRF layer. The input features of the BLSTM layer are output encodings of pre-trained BERT layers. The CRF layer is represented by lines which connect consecutive output layers. It has a state transition matrix as parameters.

Given $N$ training examples $\{x_i, y_i\}^{N}_{i=1}$, where $x_i$ is an input sequence, $y_i$ denotes a corresponding ground-truth label sequence, a negative log-likelihood objective is used as the loss function.
Thus the loss function $L$ can be defined as follow.

\begin{eqnarray}\label{logll}
\mathcal{L}=-\sum\limits_{i=1}^{N} \mathrm{log} \; p(y_i|x_i)
\end{eqnarray}
where $p(y_i|x_i)$ is the probability of the ground-truth label sequence.

At the training stage, gradient back-propagation is used to minimize the loss function. At the decoding stage, viterbi algorithm \cite{Sammut2010Viterbi} is utilized to find the most probable predicted tag sequence.

\section{Transfer task invariant knowledge from POS tagging task}

Motivated by the success of adversarial training on many tasks \cite{Goodfellow2014Generative, Ganin2017Domain, Chen2017Adversarial, Shinohara2016Adversarial, Saon2017English, Yi2018Adversarial, Yi2019LanguageCTC, Yi2019Language} ,  multi-task learning and adversarial training are combined to learn task invariant information from an extra POS tagging task. The POS tagging task is used as an auxiliary task through multi-task learning to further improve the performance of the punctuation prediction task. An adversarial loss is used to prevent the shared space from containing task specific information.

\subsection{Proposed adversarial BERT-BLSTM-CRF model}
The architecture of our proposed model is shown in Fig. \ref{fig:adv-bert-model}. The model is called adversarial BERT-BLSTM-CRF model. It consists of task shared layers, two task specific classifiers and an adversarial task discriminator.

The task shared layers are from a pre-trained BERT model, which has a stack of 12 identical layers.
The task specific classifiers are used for a punctuation predicting task and a POS tagging task, respectively. Both of them consist of BLSTM-CRF layers.

For the punctuation predicting task, one example of inputs and outputs are listed in Table \ref{tab:example-pun}. The inputs of the BERT layers are words, e.g. ``\textit{Susan where is the national library}'', while the outputs of the punctuation prediction task are punctuation marks, such as ``\textit{COMMA O O O O QUESTION}''. More details of punctuation marks are presented in Section \ref{exp:iwslt}.

For the POS tagging task, an example of inputs and outputs are listed in Table \ref{tab:example-pos}. the inputs of the BERT layers are words, e.g. ``\textit{Oh it is a beautiful morning}'', while the outputs of the POS tagging task are POS tags, e.g. ``\textit{UH PRP VBZ DT JJ NN}''. More details of POS tags are introduced in Section \ref{exp:ptb}.

However, the shared layers may learn some unnecessary task specific information.
Adversarial strategy is used to prevent the shared parameter from learning task dependent information. This idea is implemented by an adversarial task discriminator.

The adversarial task discriminator is used to recognize the task label of each sequence using the task shared features.
The outputs of the shared layers are converted into a fixed-size real-valued vector by a max-pooling layer.
The fixed-size vector is the input of the task discriminator through a gradient reversal layer (GRL) \cite{Ganin2017Domain, Ganin2015Unsupervised}. The task discriminator is implemented as a fully connected (FC) neural network with a single hidden layer. Rectified linear units (ReLU) \cite{Andrew2013Relu} are used as the activation functions of the hidden layer.

The GRL is introduced to ensure that the feature distributions are as indistinguishable as possible for the task discriminator.
Therefore, the adversarial BERT-BLSTM-CRF is to learn a representation that can generalize well from one task to another. They ensure that the internal representation of the shared parameters contains no task discriminative information.

\subsection{Multi-task learning}

\begin{table*}
\caption{Overall data distributions of IWSLT datasets for the punctuation predicting task.}
\label{tab:dataset-punc}
\centering
\begin{tabular}{cc cccccc}
\hline
{Dataset} &
{\#TED Talks} &
{\#Sentences} &
{\#Tokens} &
{\#\textit{COMMA}} &
{\#\textit{PERIOD}} &
{\#\textit{QUESTION}} &
{\#\textit{O}}\\
\hline
{Training} & 1,690 & 143,991 & 2,102,417 & 158,499 & 132,680 & 11,311 & 1,799,927\\
{Development} & 20 & 20,635 & 295,800 & 22,475 & 1,8940 & 1,695 & 252,690\\
{Test (\textit{Ref.})} & 8 & 861 & 12,626 & 830 & 808 & 53 & 10,935\\
{Test (\textit{ASR})} & 8 & 852 & 12,822 & 798 & 810 & 42 & 11,172\\
\hline
\end{tabular}
\end{table*}

Multi-task learning is one instance of transfer learning \cite{Caruana1997Multi}. The model is trained simultaneously on the training data of multiple tasks. Each task has its own private layers to estimate the posterior probabilities of task specific labels.

For the $m$-th task, given a dataset with $N_m$ training samples $\{x_i^{(m)}, y_i^{(m)}\}^{N_m}_{i=1}$, where $\{x_i^{(m)}, y_i^{(m)}\}$ is the $i$-th training sample,
$x_i^{(m)}$ is an input sequence, $y_i^{(m)} \in \{1, ..., S_y^{(m)}\}$ is the corresponding labels for the input sequence, $S_y^{(m)}$ is the total number of labels. The multi-task model is trained to minimize the negative log-likelihood for all the tasks. So the loss function of the multi-task model is defined as:
\begin{eqnarray}\label{ce-mul}
\mathcal{L}_{\mathrm{tasks}}=-\sum\limits_{m=1}^M\sum\limits_{i=1}^{N_m} \mathrm{log} \; p(y_i^{(m)}|x_i^{(m)})
\end{eqnarray}
where $m$ denotes the index of the $m$-th task, $M$ is the total number of the tasks, $p(y_i^{(m)}|x_i^{(m)})$ is computed with a parametric classifier.

In this paper, we only use two tasks. So $M$ is equal to 2. The two tasks are a punctuation predicting task and a POS tagging task. The punctuation predicting task is defined as \textit{PUN}. The POS tagging task is denoted by \textit{POS}. So the output labels of the task classifier are \textit{PUN} and \textit{POS}.

\subsection{Adversarial training}
A task discriminator is used to recognize the task label during adversarial training stage.
The gradients minimizing task classification errors are passed back with an opposite sign to the shared layers through the GRL. Thus it ensures the feature distributions over all tasks are as indistinguishable as possible for the task discriminator.

Given an additional task label for each training sample $\{x_i^{(m)}, y_i^{(m)}, m\}$, where $m \in \{ 1, ..., M\}$ denotes the task label for each sequence, and $M$ is the total number of tasks. The loss function of the task discriminator is formulated as:
\begin{eqnarray}\label{ce-adv}
\mathcal{L}_{\mathrm{adv}}=-\sum\limits_{m=1}^M\sum\limits_{i=1}^{N_m} \mathrm{log} \; p(m|x_i^{(m)})
\end{eqnarray}

Although the task classifier is optimized to minimize the task classification error, the gradient of the task classifier is negative so that the bottom shared parameters are trained to be task invariant.

\subsection{The final loss function}
Adversarial multi-task learning is used to jointly optimize the two loss functions: $\mathcal{L}_{\mathrm{tasks}}$ and $\mathcal{L}_{\mathrm{adv}}$.
For the standard multi-task learning, the shared representations are optimized in order to minimize the loss of the primary and auxiliary task. Adversarial multi-task learning is different from the standard multi-task learning.
For adversarial multi-task learning, the shared parameters are trained to maximize both the classification accuracies of the punctuation predicting task and the POS tagging task, but to minimize the classification accuracies of the task discriminator.
However, the adversarial multi-task learning works adversarially to the task discriminator through GRL. It encourages task independent features to emerge in the course of the optimization. So the shared features become punctuation marks and POS tags discriminative but task invariant. The improved task invariance leads to the improved performance of the punctuation prediction task. So the final loss function of adversarial multi-task learning is defined as:
\begin{eqnarray}\label{ce-mul-adv}
\mathcal{L}_{\mathrm{total}}=\mathcal{L}_{\mathrm{tasks}} + \lambda \mathcal{L}_{\mathrm{adv}}
\end{eqnarray}
where $\lambda \in R$ is the loss weight, $\lambda$ is gradually increased from 0 to 1 as epoch increases so that the model is stably trained \cite{Ganin2017Domain}.

There are no parameters associated with the GRL. The GRL acts as an identity transformation during the feed-forward. However, at the back-propagation stage, the GRL takes the gradient from the subsequent level and changes its sign, such as multiplying by -1, before passing it to the preceding layer. That means the gradient is reversed through the GRL by multiplying -$\lambda$. Thus the shared layers can learn task invariant knowledge from the POS tagging task.

\subsection{Decoding}
At the decoding stage, the POS tagging task and the adversarial task classifier are removed, but only the punctuation prediction task is utilized to predict marks. Therefore, the model can utilize task invariant syntactic features from the POS tagging task without increasing more computation and introducing extra errors from the POS tagger. Viterbi algorithm \cite{Sammut2010Viterbi} is also used to find the most probable punctuation sequence.

\section{Experiments}

A series of experiments are conducted to evaluate the proposed method in this section. Our experiments are conducted on English IWSLT \cite{Che2016Punc} and Penn TreeBank (PTB) datasets\footnote{https://catalog.ldc.upenn.edu/LDC99T42}. IWSLT datasets are used for punctuation prediction tasks. PTB datasets are utilized for POS tagging tasks. The results are reported on two test sets of IWSLT datasets.

\begin{table}
\caption{Overall data distributions of PTB datasets for the POS tagging task.}
\label{tab:dataset-pos}
\centering
\begin{tabular}{ccc}
\hline
{Dataset} &
{\#Sentences} &
{\#Tokens}
\\
\hline
{Training} & 39,831 & 950,011 \\
{Development} & 1,699 & 40,068 \\
{Test} & 2,415 & 56,671\\
\hline
\end{tabular}
\end{table}

\begin{table*}
\caption{ Transferring parameters from pre-trained BERT to punctuation predicting models. The results of punctuation predicting models in terms of $P(\%)$ ,$R(\%)$ , $F_1(\%)$ on test sets of IWSLT2011 datasets.}
\label{tab:result-bert}
\centering
\begin{tabular}{c|l|l|cccccccccccc}
\hline
\multirow{2}{*}{Test} &
\multirow{2}{*}{Model} &
\multirow{2}{*}{Initialization} &
\multicolumn{3}{|c}{\textit{COMMA}} &
\multicolumn{3}{c}{\textit{PERIOD}} &
\multicolumn{3}{c}{\textit{QUESTION}} &
\multicolumn{3}{c}{\textit{Overall}}\\
\cline{4-15}
 & & & $P$ & $R$ & $F_1$ & $P$ & $R$ & $F_1$ & $P$ & $R$ & $F_1$ & $P$ & $R$ & $F_1$ \\
 \hline
 \multirow{4}{*}{\textit{Ref.}} & BERT-CRF & Random & 61.1 & 59.6 & 60.3 & 72.1 & 70.7 & 71.4 & 71.3 & 60.2 & 65.3 & 68.2 & 63.5 & 65.7  \\
 & BERT-CRF  & Pre-trained BERT & 73.7 & 69.1 & 71.3 & 83.7 & 78.8 & 81.2 & 75.1 & 70.1 & 72.5 & 77.5 & 72.7 & 75.0 \\
 & BERT-BLSTM-CRF  & Random & 61.9 & 59.9 & 60.9 & 72.4 & 71.1 & 71.7 & 71.5 & 61.0 & 65.8 & 68.6 & 64.0 & 66.2 \\
 & BERT-BLSTM-CRF  & Pre-trained BERT & \textbf{74.2} & \textbf{69.7} & \textbf{71.9} & \textbf{84.6} & \textbf{79.2} & \textbf{81.8} & \textbf{76.0} & \textbf{70.4} & \textbf{73.1} & \textbf{78.3} & \textbf{73.1} & \textbf{75.6}  \\
\hline
\hline
 \multirow{4}{*}{\textit{ASR}}  & BERT-CRF & Random & 56.1 & 57.1 & 56.6 & 69.1 & 71.1 & 70.1 & 64.0 & 53.6 & 58.3 & 63.1 & 60.6 & 61.7 \\
& BERT-CRF & Pre-trained BERT & 70.2 & 67.5 & 68.8 & 76.6 & 77.1 & 76.8 & 67.5 & 65.7 & 66.6 & 71.4 & 70.1 & 70.8 \\
 & BERT-BLSTM-CRF  & Random & 56.3 & 57.4 & 56.8 & 69.4 & 71.2 & 70.3 & 64.3 & 54.1 & 58.8 & 63.3 & 60.9 & 62.0 \\
& BERT-BLSTM-CRF & Pre-trained BERT & \textbf{70.7} & \textbf{68.1} & \textbf{69.4} & \textbf{77.6} & \textbf{77.5} & \textbf{77.5} & \textbf{68.4} & \textbf{66.0} & \textbf{67.2} & \textbf{72.2} & \textbf{70.5} & \textbf{71.4} \\
\hline
\end{tabular}
\end{table*}

\subsection{IWSLT datasets}
\label{exp:iwslt}

IWSLT datasets are from TED Talks, which are reorganized for predicting punctuation marks by Che et al. \cite{Che2016Punc}. There are three kinds of datasets: training set, development set and test set.

The training and development sets are provided by the training data of IWSLT2012 machine translation track, which consists of 1,710 TED Talks. Che et al. \cite{Che2016Punc} further split it into training and development sets according to the ID of TED talks. The two test sets are \textit{Ref.} and \textit{ASR}, which provided by the test data of IWSLT2011 ASR track. \textit{Ref.} is from manual transcripts of audio files. \textit{ASR} is from transcripts of the ASR system. More details of the datasets can be found in \cite{Che2016Punc}.

The datasets have four kinds of labels: \textit{O}, \textit{COMMA}, \textit{PERIOD} and \textit{QUESTION}.
\textit{O} denotes a non-punctuation mark. \textit{COMMA} denotes the kind of colons or dashes.
Exclamation marks or semicolons are denoted by \textit{PERIOD}. \textit{QUESTION} is the kind of question marks.
Table \ref{tab:dataset-punc} describes data statistics of IWSLT datasets.

\subsection{PTB datasets}
\label{exp:ptb}
PTB datasets consist of three annotation schemes: POS tagging, syntactic bracketing, and disfluency annotation.
We only use PTB POS tagging datasets in our experiments.

The PTB tagset is based on that of the Brown Corpus, but it differs from it in a number of important ways. The PTB tagset concerns the significance of syntactic context. It encodes a word¡¯s syntactic function in its POS tag whenever possible. POS assigns each word with a unique tag that indicates its syntactic role. It contains 36 POS tags\footnote{https://www.ling.upenn.edu/courses/Fall\_2003/ling001/penn\_treebank\_pos.html}, such as \textit{UH}, \textit{PRP}, \textit{VBZ}, \textit{DT}, \textit{JJ} and \textit{NN} etc.
\textit{UH} means interjection. \textit{PRP} denotes personal pronoun. \textit{VBZ} is 3rd person singular present verb. \textit{DT} means determiner. \textit{JJ} denotes adjective and \textit{NN} means singular or mass noun.
Table \ref{tab:dataset-pos} describes data statistics of PTB datasets.

\subsection{Metrics}
 All models are evaluated in terms of precision ($P$), recall ($R$), $F_1$-score ($F_1$) in our experiments. We focus on the performance of the punctuation marks. So the correctly predicted non-punctuation marks \textit{O} are ignored. We only evaluate the performance of \textit{COMMA}, \textit{PERIOD} and \textit{QUESTION} on two test sets: \textit{Ref.} and \textit{ASR}, respectively. ``\textit{Overall}'' denotes the performance of all the three punctuation marks. The results of all experiments are reported on the two test sets of IWSLT datasets: \textit{Ref.} and \textit{ASR}. More details of metrics can be found in \cite{Che2016Punc}.

\subsection{Experimental Setup}
Pre-trained BERT models are released by Google\footnote{https://github.com/google-research/bert}, implemented with the TensorFlow toolkit \cite{Abadi2016TensorFlow}. The pre-trained models include two kinds of models\footnote{https://github.com/google-research/bert\#pre-trained-models}: BERT-Large and BERT-Base.
The size of our experimental data is not large. Therefore, we use the Uncased BERT-Base model to initialize the BERT-BLSTM-CRF models. Uncased means that any case and accent markers are stripped out.

The basic architecture of the BERT-Base model is shown at the bottom of Fig. \ref{fig:bert-model} or Fig. \ref{fig:adv-bert-model}.
The encoder has a stack of $N=12$ identical layers.
The heads $h$ of the parallel self-attention are 12.
For each of these heads, we set $d_k = d_v = d_{model}/h = 64$.
Because of the reduced dimension of each head, the total computational cost is similar to that of single-head attention with full dimensionality. In order to use residual connections, we set $d_{model} = 768$.
The positional encodings have the same dimension $d_{model}$ as the embeddings layers. So the two can be sumed. The total parameters of the BERT-Base is 110M. Please see \cite{Devlin2018BERT} for pre-training details of the BERT-Base model.

The configuration of BLSTM-CRF layers is based on the work in \cite{Yi2017Distilling}, where there are one BLSTM layer and a CRF layer.
The BLSTM layer has peephole connections and a recurrent projection layer. There are two directions in the BLSTM layer: forward and backward.
Each direction is a regular LSTM layer. The LSTM layer consists of 240 memory cells and the recurrent projection layer would project the output to 120 dimensions.
We initialize the BLSTM layer by the range (-0.02, 0.02) with a uniform distribution.
Apart from clipping the activations of memory cells to range [-50, 50], the activations of other units, the weights or the estimated gradients are not limited.

The development sets are utilized for validation. If only a little improvement between two epochs on the development set has been observed, the training terminates.

\begin{table*}
\caption{Transferring knowledge from a POS tagging task to a punctuation predicting task. The results of punctuation predicting models in terms of $P(\%)$ ,$R(\%)$ , $F_1(\%)$ on test sets of IWSLT2011 datasets.}
\label{tab:result-bert-adv}
\centering
\begin{tabular}{c|l|cccccccccccc}
\hline
\multirow{2}{*}{Test} &
\multirow{2}{*}{Model} &
\multicolumn{3}{|c}{\textit{COMMA}} &
\multicolumn{3}{c}{\textit{PERIOD}} &
\multicolumn{3}{c}{\textit{QUESTION}} &
\multicolumn{3}{c}{\textit{Overall}}\\
\cline{3-14}
 & & $P$ & $R$ & $F_1$ & $P$ & $R$ & $F_1$ & $P$ & $R$ & $F_1$ & $P$ & $R$ & $F_1$ \\
\hline
\multirow{4}{*}{\textit{Ref.}} & BERT-BLSTM-CRF (Pre-trained BERT) & 74.2 & 69.7 & 71.9 & 84.6 & 79.2 & 81.8 & 76.0 & 70.4 & 73.1 & 78.3 & 73.1 & 75.6 \\
& + POS tagging task & 75.0 & 70.3 & 72.6 & 85.5 & 79.7 & 82.5 & 76.9 & 71.3 & 74.0 & 79.1 & 73.8 & 76.4 \\
& + Task classifer & 75.1 & 70.3 & 72.6 & 85.9 & 79.9 & 82.8 & 77.0 & 71.5 & 74.1 & 79.3 & 73.9 & 76.5 \\
& + Adversarial & \textbf{76.2} & \textbf{71.2} & \textbf{73.6} & \textbf{87.3} & \textbf{81.1} & \textbf{84.1} & \textbf{79.1} & \textbf{72.7} & \textbf{75.8} & \textbf{80.9} & \textbf{75.0} & \textbf{77.8}  \\
\hline
\hline
\multirow{4}{*}{\textit{ASR}} & BERT-BLSTM-CRF (Pre-trained BERT) & 70.7 & 68.1 & 69.4 & 77.6 & 77.5 & 77.5 & 68.4 & 66.0 & 67.2 & 72.2 & 70.5 & 71.4 \\
& + POS tagging task & 71.4 & 68.6 & 70.0 & 78.4 & 77.9 & 78.1 & 69.2 & 66.8 & 68.0 & 73.0 & 71.1 & 72.0 \\
& + Task classifer & 71.5 & 68.6 & 70.0 & 78.8 & 78.1 & 78.4 & 69.3 & 67.0 & 68.1 & 73.2 & 71.2 & 72.2 \\
& + Adversarial & \textbf{72.4} & \textbf{69.3} & \textbf{70.8} & \textbf{80.0} & \textbf{79.1} & \textbf{79.5} & \textbf{71.2} & \textbf{68.0} & \textbf{69.6} & \textbf{74.5} & \textbf{72.1} & \textbf{73.3}  \\
\hline
\end{tabular}
\end{table*}

\begin{table*}
\caption{Compared with other models on IWSLT2011 datasets. The results of punctuation predicting models in terms of $P(\%)$ ,$R(\%)$ , $F_1(\%)$ on test sets .}
\label{tab:result-others}
\centering
\begin{tabular}{c|l|l|cccccccccccc}
\hline
\multirow{2}{*}{Test} &
\multirow{2}{*}{Model} &
\multirow{2}{*}{Transferred Info.} &
\multicolumn{3}{|c}{\textit{COMMA}} &
\multicolumn{3}{c}{\textit{PERIOD}} &
\multicolumn{3}{c}{\textit{QUESTION}} &
\multicolumn{3}{c}{\textit{Overall}}\\
\cline{4-15}
 & & & $P$ & $R$ & $F_1$ & $P$ & $R$ & $F_1$ & $P$ & $R$ & $F_1$ & $P$ & $R$ & $F_1$ \\
 \hline
 \multirow{10}{*}{\textit{Ref.}} & CRF Best \cite{Ueffing2013Improved} & Lexical \& Syntactic & - & - & - & - & - & - & - & - & - & 49.8 & 58.0 & 53.5 \\
& DNN-A \cite{Che2016Punc} & Word vectors & 48.6 & 42.4 & 45.3 & 59.7 & 68.3 & 63.7 & - & - & - & 54.8 & 53.6 & 54.2 \\
& CNN-2A \cite{Che2016Punc} & Word vectors & 48.1 & 44.5 & 46.2 & 57.6 & 69.0 & 62.8 & - & - & - & 53.4 & 55.0 & 54.2 \\
& T-LSTM \cite{Tilk2015LSTM} & One-hot vectors & 49.6 & 41.4 & 45.1 & 60.2 & 53.4 & 56.6 & 57.1 & 43.5 & 49.4 & 55.0 & 47.2 & 50.8 \\
& T-BRNN-pre \cite{Tilk2016BRNN}  & Word vectors & 65.5 & 47.1 & 54.8 & 73.3 & 72.5 & 72.9 & 70.7 & 63.0 & 66.7 & 70.0 & 59.7 & 64.4 \\
& BLSTM-CRF \cite{Yi2017Distilling} & Word vectors & 58.9 & 59.1 & 59.0 & 68.9 & 72.1 & 70.5 & 71.8 & 60.6 & 65.7 & 66.5 & 63.9 & 65.1 \\
& Teacher-Ensemble \cite{Yi2017Distilling}  & Word vectors & 66.2 & 59.9 & 62.9 & 75.1 & 73.7 & 74.4 & 72.3 & 63.8 & 67.8 & 71.2 & 65.8 & 68.4 \\
& DRNN-LWMA-pre \cite{Kim2019Deep}  & Word vectors & 62.9 & 60.8 & 61.9 & 77.3 & 73.7 & 75.5 & 69.6 & 69.6 & 69.6 & 69.9 & 67.2 & 68.6 \\
& Self-attention \cite{Yi2019Self} & Word \& Speech vectors & 67.4 & 61.1 & 64.1 & 82.5 & 77.4 & 79.9 & 80.1 & 70.2 & 74.8 & 76.7 & 69.6 & 72.9  \\
& Our best model  & BERT \& POS task & \textbf{76.2} & \textbf{71.2} & \textbf{73.6} & \textbf{87.3} & \textbf{81.1} & \textbf{84.1} & \textbf{79.1} & \textbf{72.7} & \textbf{75.8} & \textbf{80.9} & \textbf{75.0} & \textbf{77.8}  \\
\hline
\hline
 \multirow{10}{*}{\textit{ASR}} & CRF Best \cite{Ueffing2013Improved} & Lexical \& Syntactic & - & - & - & - & - & - & - & - & - & 47.8 & 54.8 & 51.0 \\
& DNN-A \cite{Che2016Punc} & Word vectors & 41.0 & 40.9 & 40.9 & 56.2 & 64.5 & 60.1 & - & - & - & 49.2 & 51.6 & 50.4 \\
& CNN-2A \cite{Che2016Punc} & Word vectors & 37.3 & 40.5 & 38.8 & 54.6 & 65.5 & 59.6 & - & - & - & 46.4 & 51.9 & 49.1 \\
& T-LSTM \cite{Tilk2015LSTM}  & One-hot vectors & 41.8 & 37.8 & 39.7 & 56.4 & 49.3 & 52.6 & 55.6 & 42.9 & 48.4 & 49.1 & 43.6 & 46.2 \\
& T-BRNN-pre \cite{Tilk2016BRNN}  & Word vectors & 59.6 & 42.9 & 49.9 & 70.7 & 72.0 & 71.4 & 60.7 & 48.6 & 54.0 & 66.0 & 57.3 & 61.4\\
& BLSTM-CRF \cite{Yi2017Distilling}  & Word vectors & 55.7 & 56.8 & 56.2 & 68.7 & 71.5 & 70.1 & 63.8 & 53.4 & 58.1 & 62.7 & 60.6 & 61.5\\
& Teacher-Ensemble \cite{Yi2017Distilling} & Word vectors & 60.6 & 58.3 & 59.4 & 71.7 & 72.9 & 72.3 & 66.2 & 55.8 & 60.6 & 66.2 & 62.3 & 64.1\\
& DRNN-LWMA-pre \cite{Kim2019Deep}  & Word vectors & - & - & - & - & - & - & - & - & - & - & - & - \\
& Self-attention \cite{Yi2019Self}  & Word \& Speech vectors & 64.0 & 59.6 & 61.7 & 75.5 & 75.8 & 75.6 & 72.6 & 65.9 & 69.1 & 70.7 & 67.1 & 68.8 \\
& Our best model & BERT \& POS task & \textbf{72.4} & \textbf{69.3} & \textbf{70.8} & \textbf{80.0} & \textbf{79.1} & \textbf{79.5} & \textbf{71.2} & \textbf{68.0} & \textbf{69.6} & \textbf{74.5} & \textbf{72.1} & \textbf{73.3} \\
\hline
\end{tabular}
\end{table*}

\subsection{Transferring parameters from pre-trained BERT}

In this section, we evaluate the effectiveness of transferring parameters from pre-trained BERT on IWSLT datasets. Two model architectures are designed for punctuation predicting task: BERT-CRF and BERT-BLSTM-CRF. The output labels of the classification layer are three punctuation marks and one non-punctuation mark \textit{O}.

\emph{BERT-CRF}: A simple classification layer is added to the BERT layers. The classification layer is a CRF layer.

\emph{BERT-BLSTM-CRF}: We add two layers at the top of the BERT layers as shown in Fig. \ref{fig:bert-model}. They are BLSTM-CRF layers, which consist of a one-layer BLSTM and a CRF layer. A linear transformation is used to convert the 768-dimensional activations to 120-dimensional BLSTM layer.

In the first group of experiments, all the parameters of BERT-CRF and BERT-BLSTM-CRF are initialized randomly. The models are trained for 6 epochs over the training data. The Adam algorithm \cite{Kingma2015Adam} with gradient clipping and warmup is used for optimization. The $warmup\_steps$ is set to 4,000. The $batch\_size$ is set to 32, which means that each batch contains 32 sentences. The rate of dropout is set to 0.1. The initial learning rate is 5e-4. The learning rate is varied over the course of training, according to the formula in \cite{Vaswani2017Attention}.

In the second group of experiments, the parameters of BERT layers of BERT-CRF and BERT-BLSTM-CRF are first initialized with the parameters of the pre-trained BERT model. Then all of the parameters are jointly fine-tuned using training data for the punctuation predicting task. The models are fine-tuned for 3 epochs over the training data. The $batch\_size$ is set to 32. The rate of dropout is set to 0.1. we select the best fine-tuning learning rate of 5e-5 on the development set.

The results of the models are reported in Table \ref{tab:result-bert}.
The results show that the BERT-BLSTM-CRF models obtain better results than the BERT-CRF models accordingly. The BERT-CRF models with random initialization achieve the worst results among our models on both two test sets. But the BERT-BLSTM-CRF models with transferred parameters from pre-trained BERT model obtain the best performance on both two test sets: \textit{Ref.} and \textit{ASR}.

The results also show that the models initialized from pre-trained BERT model outperform the models with random initialization significantly.
The BERT-CRF model initialized from pre-trained BERT model obtains better performance than that with random initialization by 9.3\% and 9.1\% absolute overall $F_1$-score on \textit{Ref.} and \textit{ASR} test sets, respectively.
The BERT-BLSTM-CRF model initialized from pre-trained BERT model outperforms that initialized randomly by 9.4\% and 9.4\% absolute overall $F_1$-score on \textit{Ref.} and \textit{ASR} test sets, respectively.

In the rest of our experiments, we use BLSTM-CRF layers as task specific classification layers on top of the BERT layers for both punctuation predicting task and POS tagging task. In addition, the BERT layers are initialized by the parameters of the pre-trained BERT model.

\subsection{Transferring knowledge from POS tagging task}

A series of experiments are performed to evaluate the performance of the punctuation predicting task trained with the knowledge transferred from a POS tagging task. The POS tagging task is trained on PTB POS tagging datasets. We fine-tune the models for 3 epochs with a batch size of 32.

In the first group of experiments, we train the punctuation restoring task only with the help of the POS tagging task via multi-task learning. The architecture of the POS tagging specific layers is identical to that of the punctuation restoring specific layers. It consists of one-layer BLSTM and a CRF layer. The two tasks share pre-trained BERT layers. The output of the BERT layers are the input of the two tasks. A linear transformation is used to convert the 768-dimensional output to 120-dimensional input for the two task specific BLSTM layers. The output labels of the POS tagging task are 36 POS tags.

In the second group of experiments, we add a task discriminator to the aforementioned multi-task model. The task discriminator has one max-pooling layer, a FC layer and a softmax layer. The ReLU activation functions are used in the 1024-dimensional FC layer. The task classifier has two output labels: \textit{PUN} and \textit{POS}. \textit{PUN} denotes the punctuation predicting task. \textit{POS} is referred as to the POS tagging task. During training, we select a task from \{\textit{PUN}, \textit{POS}\} at each iteration. Then, we use a batch of training samples from the given task to update the parameters. The Adam algorithm \cite{Kingma2015Adam} with gradient clipping and warmup is used for to optimize the loss function. The punctuation predicting task and
POS tagging task may have different convergence rate. So we repeat the above iterations until early stopping according to the punctuation predicting performance.

In the third group of experiments, we further add a GRL in the above-mentioned task discriminator. The GRL is after the max-pooling layer and before the FC layer as shown in Fig. \ref{fig:adv-bert-model}. The GRL has no parameters.
The dropout rate is fixed at 0.1.
The loss weight $\lambda$  is initiated at 0 and is gradually changed to 1 with the following formula \cite{Ganin2017Domain}:
\begin{eqnarray}\label{update-lambda}
&& \lambda=\frac{2}{1+exp(-\gamma \cdot p)}-1
\end{eqnarray}
where $p$ is the training progress linearly changing from 0 to 1, $\gamma$ is set to 10 in all experiments.

This strategy allows the task classifier to be less sensitive to noisy signal at the early stages of the training procedure. Note that the $\lambda$ is used only for updating the shared BERT layers. However, for updating the task classification component, we use a fixed $\lambda=1$, to ensure that the latter trains as fast as the two task specific classifiers \cite{Ganin2017Domain}.
The Adam algorithm is also used for to optimize the final loss function.

The results of the above three groups of experiments are listed in Table \ref{tab:result-bert-adv}.
The results show that with the help of the POS tagging task via multi-task learning, the punctuation restoring task obtains performance improvement both on \textit{Ref.} and \textit{ASR} test sets. The results also demonstrate that when the multi-task model with an additional task classifier, the performance of the punctuation predicting models improve moderately on two test sets.
However, the punctuation predicting models achieve further obvious improvements when the multi-task model with an extra adversarial task discriminator on \textit{Ref.} and \textit{ASR} test sets, respectively.

\subsection{Compared with other methods}

We also compare our proposed models with previous models on IWSLT2011 datasets. The previous results are listed in Table \ref{tab:result-others}. "Transferred Info." in Table \ref{tab:result-others} denotes the type of transferred information from text data.

\emph{CRF best} is the best model proposed by Ueffing et al. \cite{Ueffing2013Improved}.
\emph{DNN-A} and \emph{CNN-2A} are the best models from \cite{Che2016Punc}, in which Che et al. half the value of softmax output for class ¡°O¡±.
\emph{T-LSTM} represents the first stage model from \cite{Tilk2015LSTM} that Tilk et al. train on the English IWSLT2011 dataset.
\emph{T-BRNN-pre} is the best attention model proposed by Tilk et al. \cite{Tilk2016BRNN}.
\emph{BLSTM-CRF} denotes the best single model trained in \cite{Yi2017Distilling}.
\emph{Teacher-Ensemble} is the best ensemble model proposed by Yi et al. \cite{Yi2017Distilling}.
\emph{DRNN-LWMA-pre} represents the best multi-head attention based model from \cite{Kim2019Deep}.
\emph{Self-attention} \cite{Yi2019LanguageCTC} achieves the state-of-the-art results.

CRF best, DNN-A, CNN-2A, T-LSTM, T-BRNN-pre, BLSTM-CRF, Teacher-Ensemble and DRNN-LWMA-pre models in Table \ref{tab:result-others} are trained only with text data. Whereas Self-attention model is trained using both lexical and prosody features. Our models are trained only using text data.

The results show that our best models with purely lexical features outperform all the previous state-of-the-art models.
When compared with the best model in \cite{Kim2019Deep}, the overall $F_1$-score of our best model improves absolutely by 9.2\% on \textit{Ref.} test set.
When compared with the lexical and prosody model: Self-attention \cite{Yi2019LanguageCTC}, the overall $F_1$-score of our best model also improves absolutely by 4.9\% and 4.5\% on \textit{Ref.} and \textit{ASR} test set, respectively.

\section{Discussions}

The above experimental results show that the proposed adversarial transfer learning is effective.
Some interesting observations are made as follows.

The punctuation predicting models obtain significant performance improvement via transferred parameters from pre-trained BERT model.
The parameters transferred from the pre-trained BERT model are used to initialize the punctuation predicting models. It is helpful for at least three reasons.
One reason is that the punctuation predicting model has parameters for feature types observed in the a large amount of external unlabeled text data as well as punctuated text data. Thus it has better feature coverage. The second reason is that the training objective is non-convex. So this initialization can be benefited in avoiding bad local optima. The third reason is that pre-trained BERT model is a deep bidirectional language model trained on both left and right context. Thus the punctuation predicting model can use left-to-right and right-to-left representations transferred from the pre-trained BERT model. The bidirectional knowledge is useful for predicting punctuation marks.

The punctuation predicting task benefits from a POS tagging task. The syntactic features convey useful information if the input text is formal and well-structured. POS tagging corpus encode a word¡¯s syntactic function in its POS tag whenever possible. POS assigns each word with a unique tag that indicates its syntactic role. So the punctuation predicting task can learn helpful syntactic knowledge from the POS tagging task.

The punctuation predicting models gain further obvious performance improvement with task invariant knowledge.
Although the punctuation restoring task obtains performance improvement with the help of the POS tagging task via multi-task learning, the punctuation predicting models achieve further obvious improvements when the multi-task model trained with an extra adversarial task discriminator. The main possible reason is that the shared layers of the multi-task model may learn some unnecessary task specific features. However, the adversarial loss makes the shared layers to prevent from learning the task dependent information. So the punctuation predicting models with an adversarial task classifier can learn more task invariant features. The transferred task invariant knowledge are helpful for improving the performance of the punctuation predicting model.

In summary, all the punctuation predicting models benefit from both better feature coverage and better initialization, as well as syntactic knowledge via transfer learning. Moreover, the adversarial strategy forces the shared layers to prevent from containing task dependent information. The punctuation predicting models benefit from the task invariant features by adversarial transfer learning.


\section{Conclusion}

This paper proposes adversarial transfer learning to improve the performance of punctuation predicting tasks. Bidirectional representations are transferred from a pre-trained BERT model to punctuation prediction models. Furthermore, task invariant knowledge is learnt for the punctuation prediction task with an auxiliary POS tagging task via adversarial multi-task learning.
Experiments are conducted on IWSLT2011 datasets. The results demonstrate that the punctuation predicting models with transferred parameters from pre-trained BERT model outperform the models with random initialization significantly. The results also show that the punctuation predicting models with task invariant knowledge obtain further performance improvement. Our best model outperforms the previous state-of-the-art models. Future work includes applying the proposed method to other speech signal processing tasks.
\section*{Acknowledgments}

This work is supported by the National Key Research \& Development Plan of China (No. 2017YFC0820602) and the National Natural Science Foundation of China (NSFC) (No.61425017, No. 61773379, No. 61603390, No. 61771472, No. 61901473), and Inria-CAS Joint Research Project (No. 173211KYSB20190049).

\ifCLASSOPTIONcaptionsoff
  \newpage
\fi



%

\bibliographystyle{unsrt}

\bibliography{strings,refs}


%

\begin{IEEEbiography}[{\includegraphics[width=1in,height=1.25in,clip,keepaspectratio]{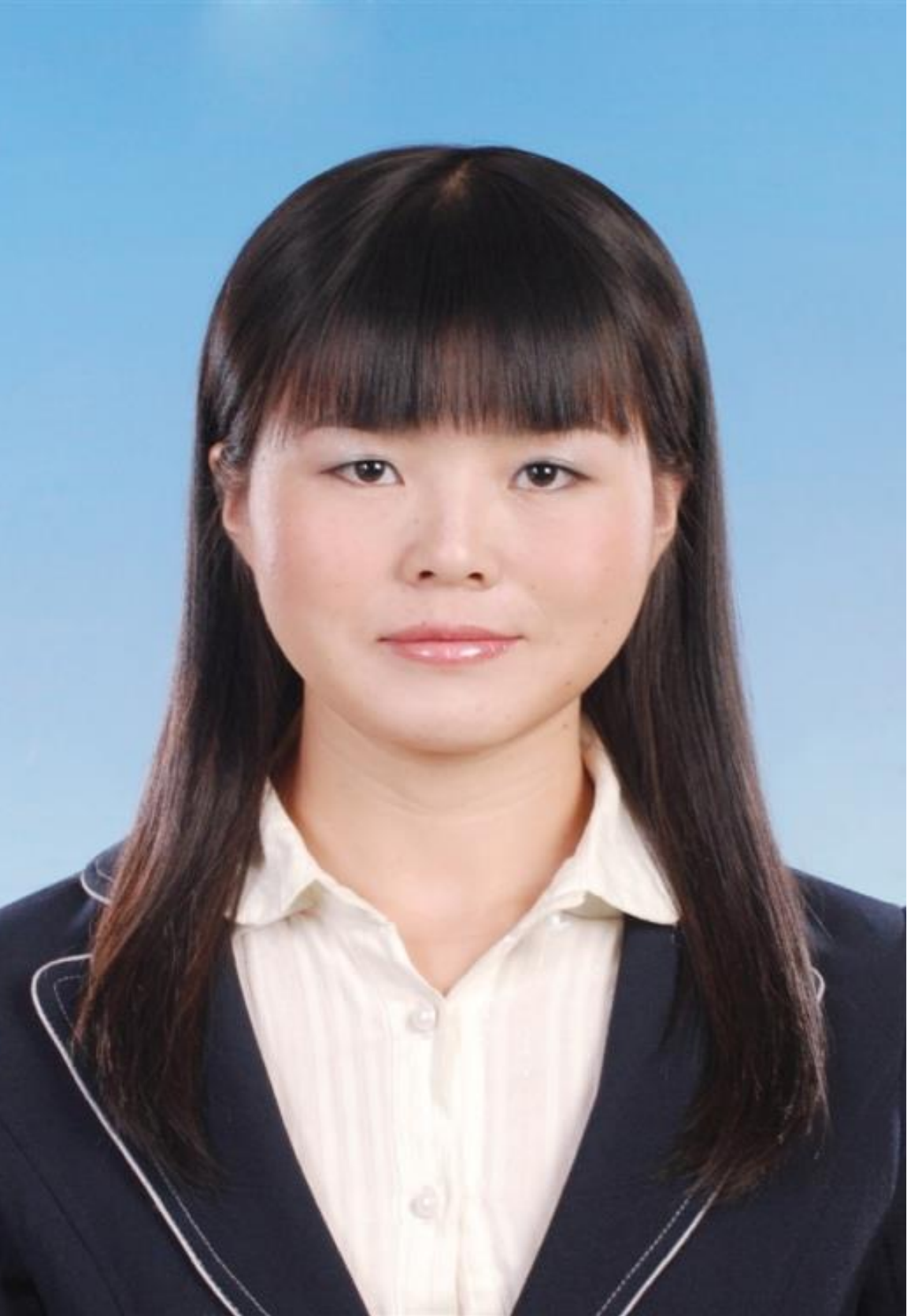}}]{Jiangyan Yi}
received the Ph.D. degree from the University of Chinese Academy of Sciences, Beijing, China, in 2018, and the M.A. degree from the Graduate School of Chinese Academy of Social Sciences, Beijing, China, in 2010. She was a Senior R\&D Engineer with Alibaba Group during 2011 to 2014. She is currently an Assistant Professor with the National Laboratory of Pattern Recognition, Institute of Automation, Chinese Academy of Sciences, Beijing, China. Her current research interests include speech processing, speech recognition, acoustic model, deep learning, and transfer learning.
\end{IEEEbiography}

\begin{IEEEbiography}[{\includegraphics[width=1in,height=1.25in,clip,keepaspectratio]{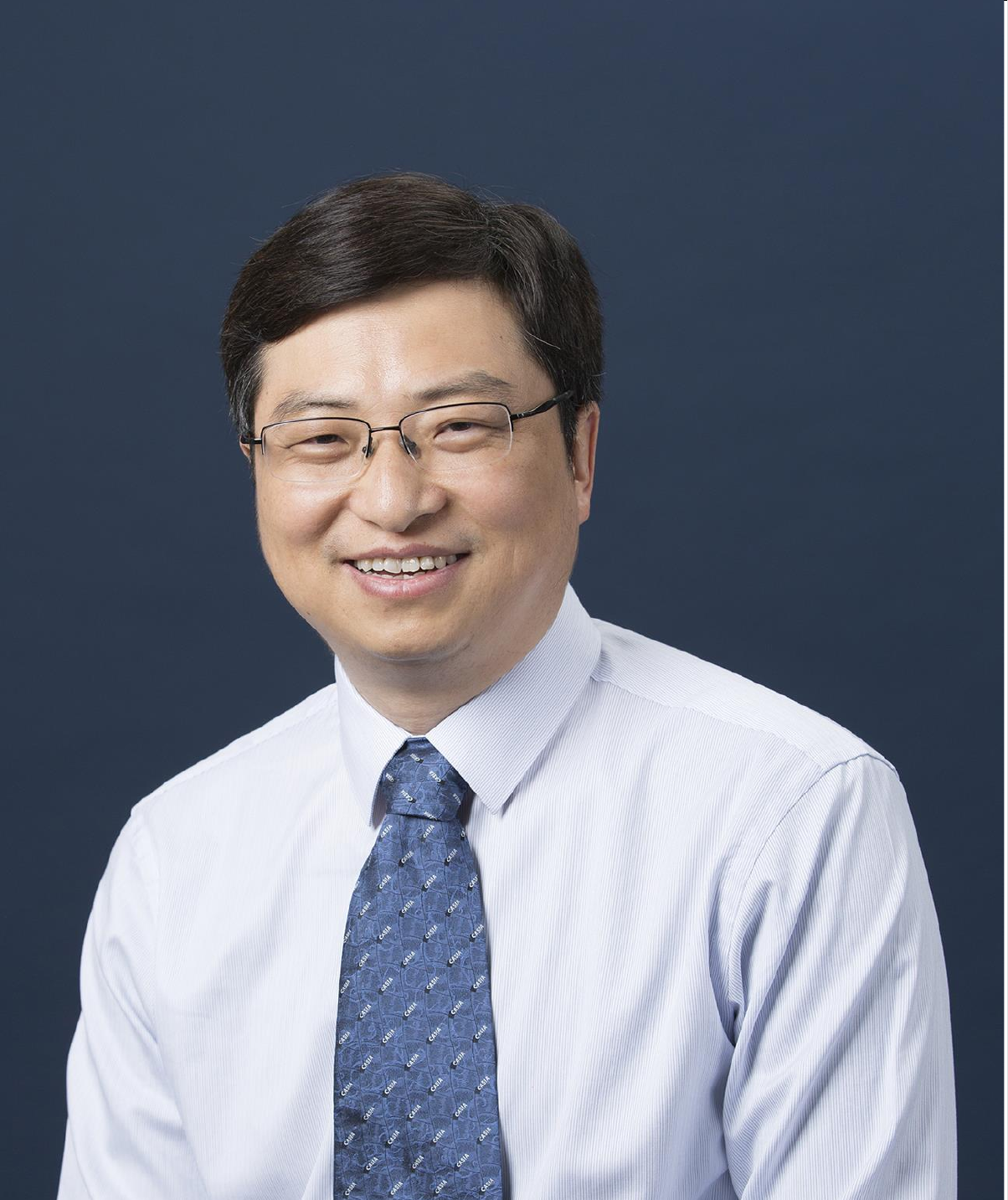}}]{Jianhua Tao}
received his Ph.D. degree from Tsinghua University, Beijing, China, in 2001, and the M.S. degree from Nanjing University, Nanjing, China, in 1996. He is currently a Professor with NLPR, Institute of Automation, Chinese Academy of Sciences, Beijing, China. He has authored or coauthored more than eighty papers on major journals and proceedings including the IEEE TRANSACTIONS ON AUDIO, SPEECH, AND LANGUAGE PROCESSING. His current research interests include speech recognition, speech synthesis and coding methods, human¨Ccomputer interaction, multimedia information processing, and pattern recognition. He is the Chair or Program Committee Member for several major conferences, including ICPR, ACII, ICMI, ISCSLP, NCMMSC, etc. He is also the Steering Committee Member for the IEEE TRANSACTIONS ON AFFECTIVE COMPUTING, an Associate Editor for Journal on Multimodal User Interface and International Journal on Synthetic Emotions, and the Deputy Editor-in-Chief for Chinese Journal of Phonetics. He was the recipient of several awards from the important conferences, such as Eurospeech, NCMMSC, etc.
\end{IEEEbiography}

\begin{IEEEbiography}[{\includegraphics[width=1in,height=1.25in,clip,keepaspectratio]{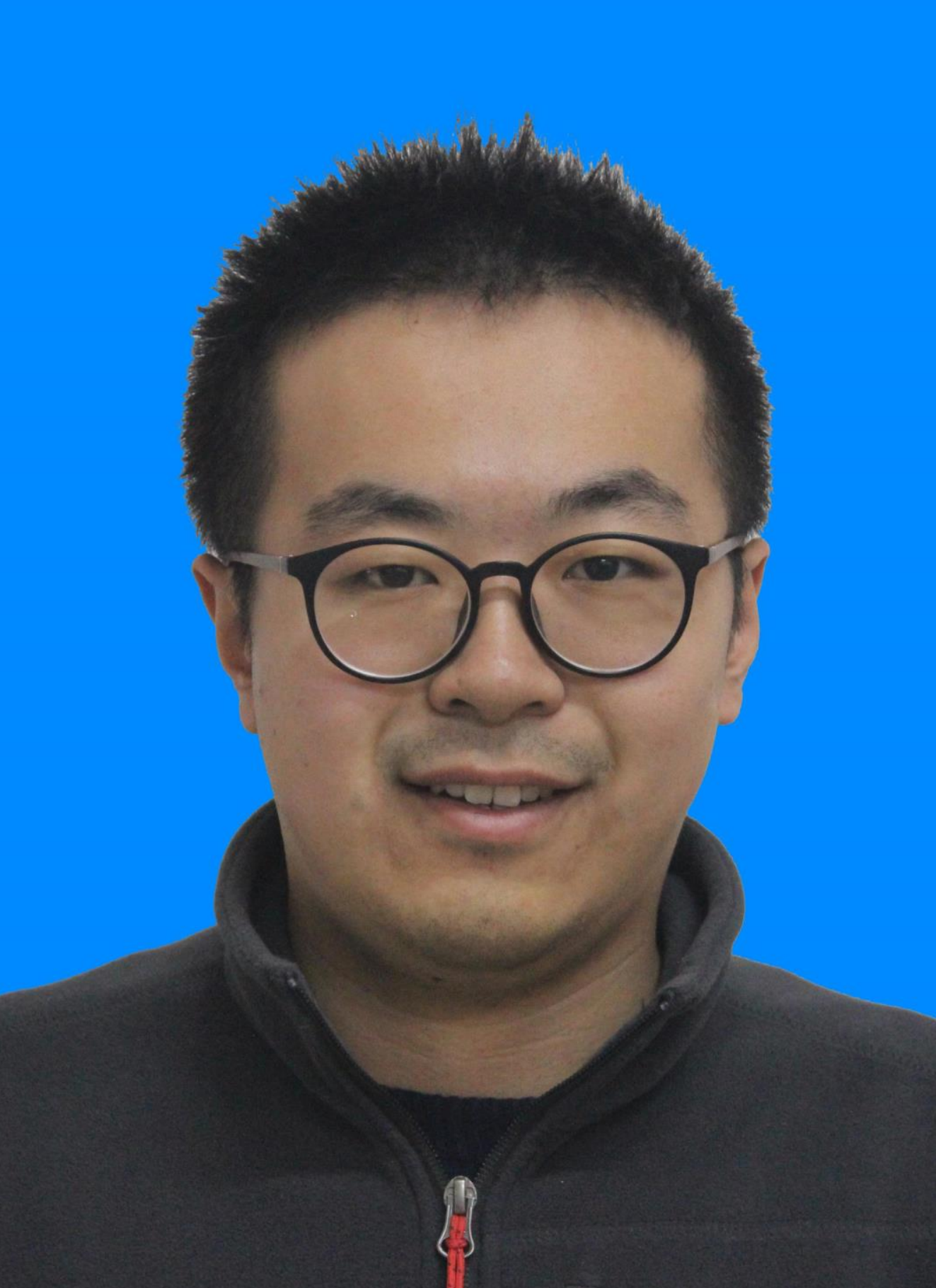}}]{Ye Bai}
received the B.S. degree from China Agricultural University, Beijing, China, in 2016. He is
currently working toward the Ph.D. degree with the University of Chinese Academy of Sciences, Beijing, China. His current research interests include speech recognition, language modeling, and keyword spotting.
\end{IEEEbiography}

\begin{IEEEbiography}[{\includegraphics[width=1in,height=1.25in,clip,keepaspectratio]{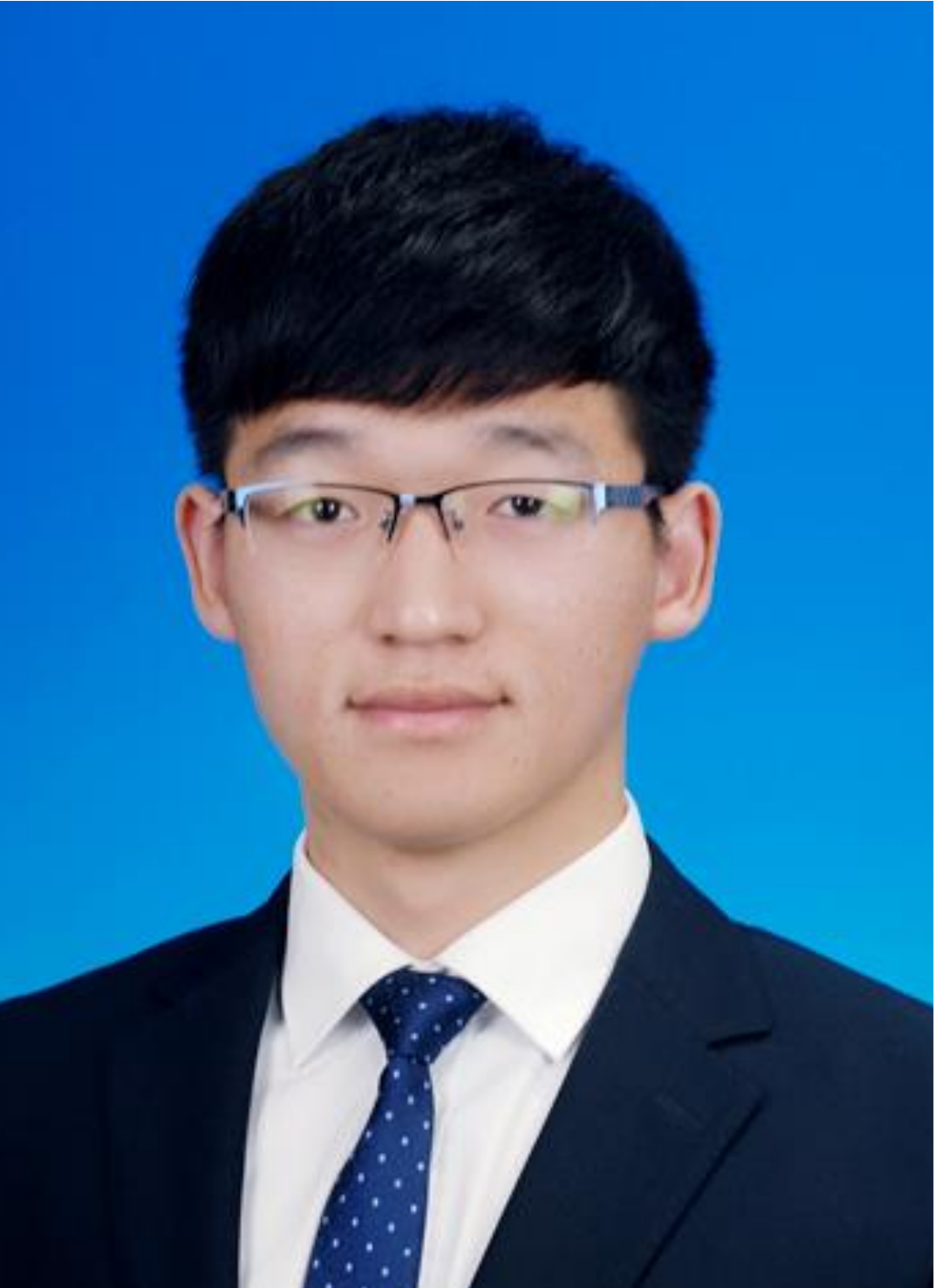}}]{Zhengkun Tian}
received the B.S. degree from Tianjin University, Tianjin, China, in 2017. He is currently working toward the Ph.D. degree with the University of Chinese Academy of Sciences, Beijing, China. His current research interests include speech recognition, speaker verification and identification.
\end{IEEEbiography}

\begin{IEEEbiography}[{\includegraphics[width=1in,height=1.25in,clip,keepaspectratio]{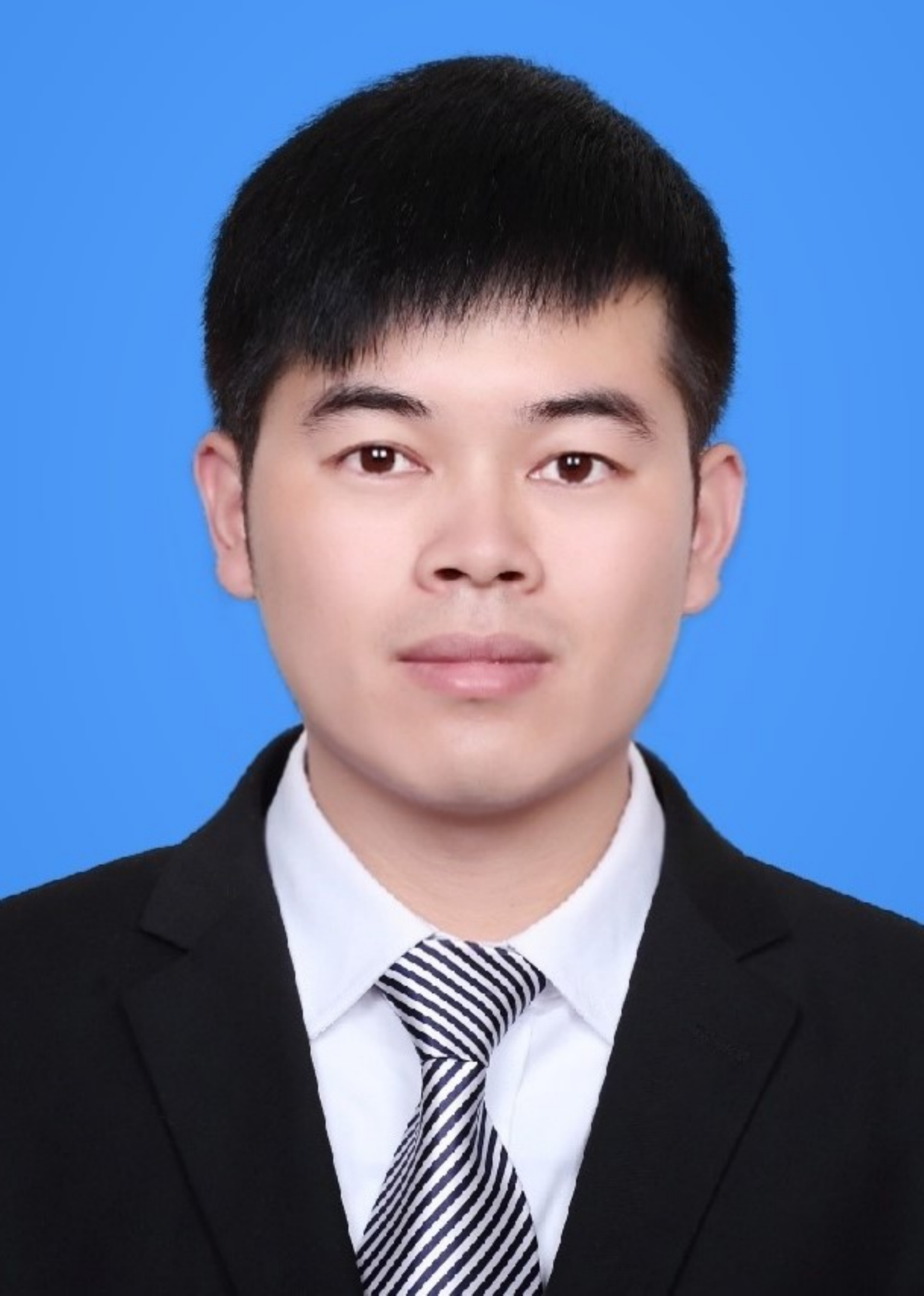}}]{Cunhang Fan}
received the B.S. degree from the Beijing University of Chemical Technology (BUCT), Beijing, China, in 2016. He is currently
working toward the Ph.D. degree with the National Laboratory of Pattern Recognition (NLPR), Institute of Automation, Chinese Academy of Sciences (CASIA), Beijing, China. His current research interests include speech separation, speech enhancement,
speech recognition and speech signal processing.
\end{IEEEbiography}




\end{document}